\documentclass[conference]{IEEEtran}
\IEEEoverridecommandlockouts
\usepackage{cite}
\usepackage{amsmath,amssymb,amsfonts}
\usepackage{algorithmic}
\usepackage{graphicx}
\usepackage{textcomp}
\usepackage{xcolor}
\usepackage{multirow}
\usepackage[misc,geometry]{ifsym}
\def\BibTeX{{\rm B\kern-.05em{\sc i\kern-.025em b}\kern-.08em
    T\kern-.1667em\lower.7ex\hbox{E}\kern-.125emX}}
\begin{document}
\title{Gaze-Assisted Human-Centric Domain Adaptation for Cardiac Ultrasound Image Segmentation
\thanks{* Corresponding author. (Email: rongjun\_ge@seu.edu.cn)

$\dagger$ These authors contributed equally to this work

This study was supported by the Natural Science Foundation of Jiangsu Province (No. BK20210291); the National Natural Science Foundation of China (No. 62101249, No. T2225025 and No. 62136004); the Jiangsu Shuangchuang Talent Program (No. JSSCBS20220202); the China Postdoctoral Science Foundation (No. 2021TQ0149 and No. 2022M721611)
}
}
\author{\IEEEauthorblockN{\textit{Ruiyi Li$^{1, \dagger}$, Yuting He$^{2, \dagger}$, Rongjun Ge$^{3, *}$, Chong Wang$^{1}$, Daoqiang Zhang$^{1}$, Yang Chen$^{2}$, Shuo Li$^{4}$}}\\
\IEEEauthorblockA{$^{1}$College of Artificial Intelligence, Nanjing University of Aeronautics and Astronautics, China\\
$^{2}$School of Computer Science and Engineering, Southeast University, China\\$^{3}$School of Instrument Science and Engineering, Southeast University, China\\
$^{4}$Department of Biomedical Engineering, Case Western Reserve University, USA
}}
\maketitle
\begin{abstract}
Domain adaptation (DA) for cardiac ultrasound image segmentation is clinically significant and valuable. However, previous domain adaptation methods are prone to be affected by the incomplete pseudo-label and low-quality target to source images. Human-centric domain adaptation has great advantages of human cognitive guidance to help model adapt to target domain and reduce reliance on labels. Doctor gaze trajectories contains a large amount of cross-domain human guidance. To leverage gaze information and human cognition for guiding domain adaptation, we propose gaze-assisted human-centric domain adaptation (GAHCDA), which reliably guides the domain adaptation of cardiac ultrasound images. GAHCDA includes following modules: (1) Gaze Augment Alignment (GAA): GAA enables the model to obtain human cognition general features to recognize segmentation target in different domain of cardiac ultrasound images like humans. (2) Gaze Balance Loss (GBL): GBL fused gaze heatmap with outputs which makes the segmentation result structurally closer to the target domain. The experimental results illustrate that our proposed framework is able to segment cardiac ultrasound images more effectively in the target domain than GAN-based methods and other self-train based methods, showing great potential in clinical application.  
\end{abstract}

\begin{figure*}[t]
\centering
\includegraphics[width=0.83\textwidth]{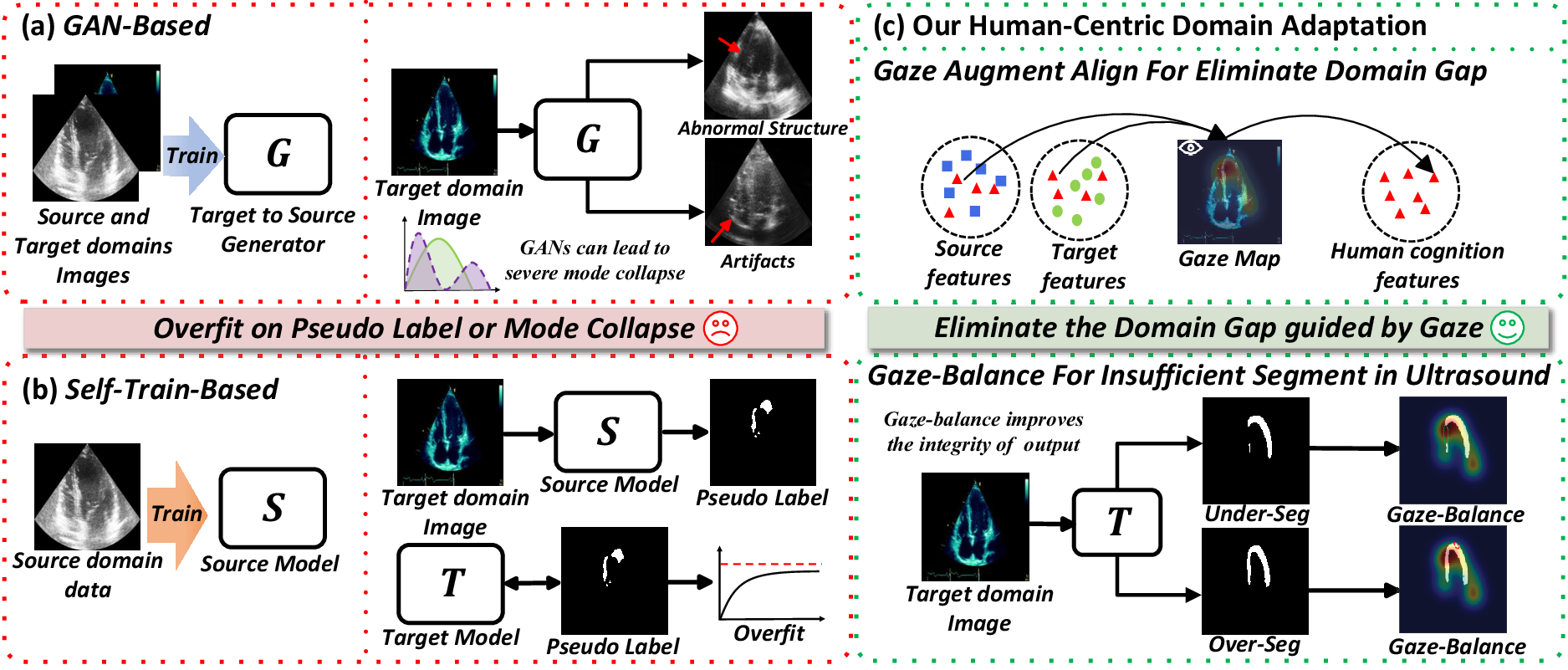}
\vspace{-0.2cm}
\caption{(a) Domain adaptation based on GAN is prone to mode collapse. (b) Domain adaptation based on self-training is prone to overfit on pseudo labels. (c) Human-centric domain adaptation uses gaze information from doctors’ diagnoses to help the model adapt to the domain and reduce overfitting.} \label{fig1}
\vspace{-0.22cm}
\end{figure*}

\section{Introduction}
Domain adaptation for cardiac ultrasound segmentation holds significant clinical value. Through domain adaptation, models are able to address the performance degradation of models across different domain segmentation. However, existing domain adaptation methods face challenges in ultrasound images. Models are prone to overfit in the target domain because they are unable to learn human cognition general features to eliminate the domain gap. Furthermore, there is a large gap in style between source domain and target domain in cardiac ultrasound images due to the differences in sampling devices, sampling settings, and sampling angles. 

Existing domain adaptation methods are limited when applied to cardiac ultrasound images. Fully supervised domain adaptation methods require additional annotations\cite{vesal2022domain,liu2024imaging}. Semi-supervised or unsupervised domain adaptation methods adapt to different domains using GAN or self-training approaches\cite{semi_chen2023deep,semi_xu2021shadow,semi_yang2021contrastive_semi,guSemi2023}. As shown in Fig.\ref{fig1} (a), GAN-based domain adaptation methods struggle to adapt well to the gap in different domains of cardiac ultrasound images\cite{chen2019synergistic,chen2020unsupervised,CycleGAN2017,choi2019self,yan2019domain,iacono2023structure,xie2020mi,yao2022novel,dong2020can,tomar2021self,pan2020unsupervised}, leading to mode collapse\cite{bau2019seeing_Gan}. As shown in Fig.\ref{fig1} (b), domain adaptation based on self-train lacks human cognition guidance, making it prone to overfit on source domain data, resulting in overfit on pseudo-labels and incomplete segmentation \cite{frequency,ASC,yang2020fda,yang2022source,wu2021unsupervised,guo2021metacorrection,zou2018unsupervised,bolte2019unsupervised,kamnitsas2017unsupervised,zhang2019category,hoyer2022daformer}. 

Human-centric domain adaptation holds great advantages of cognitive guidance in cardiac ultrasound domain adaptation. Doctor gaze trajectories contains a large number of cross-domain human guidance. By recording gaze trajectories using an eye tracker, we are able to extract the doctor's cross-domain recognition knowledge. As shown in Fig.\ref{fig1} (c), utilizing this cross-domain recognition knowledge helps the model eliminate the domain gap between source and target domain.

In this paper, we propose gaze-assisted human-centric domain adaptation (GAHCDA) framework for cardiac ultrasound image segmentation tasks. GAHCDA includes following modules: (1) Gaze Aug Align Module (GAA): This is a feature alignment module that integrates human cognition from gaze heatmaps. It utilizes a cross-attention\cite{dosovitskiy2020ViT} to fuse the encoder features of teacher with gaze heatmap and aligns the features of the student with fused human cognition features, thereby extracting human cognition general features between the source and target domains. (2) Gaze Balance Loss (GBL): By using gaze heatmaps, the model loss is more focused on gaze area, allowing the model to avoid over/under-segmentation in the gaze area rather than being restricted to pseudo-label.

\section{Method}
\subsection{GAHCDA Framwork}
In the context of human-centric domain adaptation for cardiac ultrasound image segmentation, we have a source domain dataset ${D}_{s}=\{{X}_{s},{Y}_{s}\}_{s=1}^{M}$ and a target domain dataset ${D}_{t}=\{{X}_{t}\}_{t=1}^{N}$, where $X$ represents images and $Y$ represents labels. The objective of GAHCDA is to train a teacher model using source data and use the teacher model to obtain pseudo-labels $\hat{Y}_{t}^T$ for target domain images ${X}_{t}$. These pseudo-labels are then used to train a student model for the target domain data. As shown in Fig. \ref{Methods}, our GAHCDA includes the gaze augment align module(GAA), which helps the student model quickly extract human cognition general features. It also includes gaze balance loss (GBL), which utilizes gaze heatmaps to address the over/under-segmentation problem.
\subsection{Gaze Augment Align Module}
\begin{figure*}[t]
\centering
\includegraphics[width=0.83\textwidth]{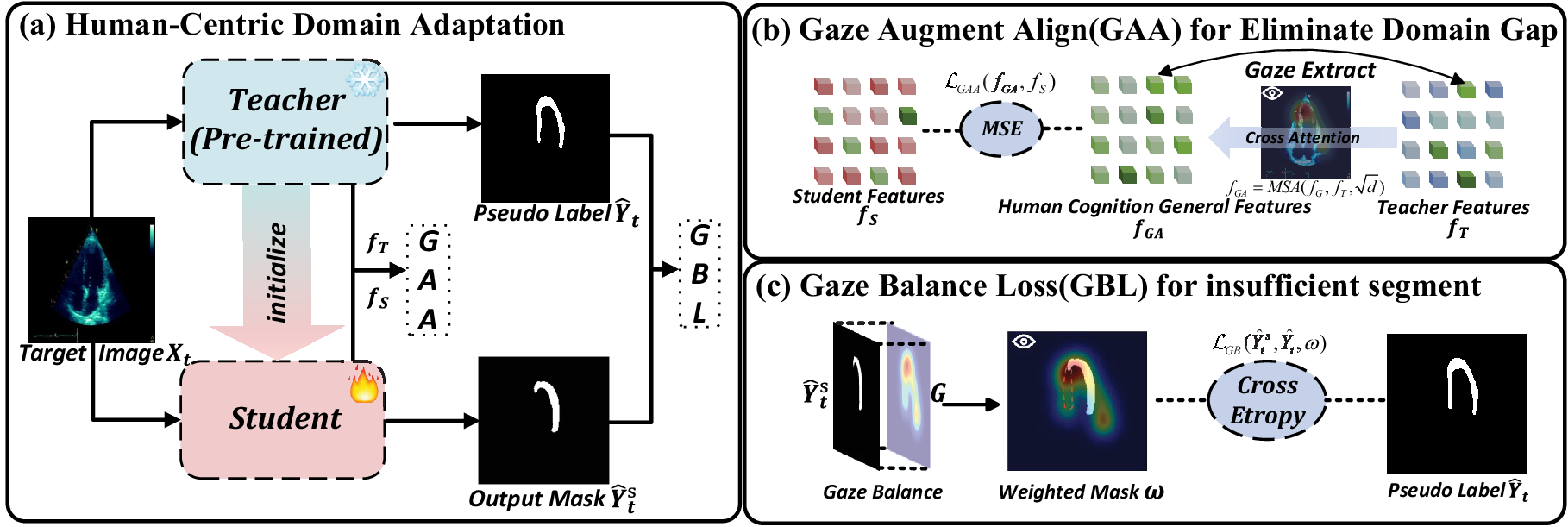}
\vspace{-0.25cm}
\caption{The Gaze-Assisted Human-Centric domain adaptation method uses gaze guidance to help student models acquire human-like cross-domain segmentation and recognition capabilities. (a) Teacher model parameters trained on the source domain are used to initialize the student model in the target domain. (b) The Gaze Augment Align Module (GAA) module uses gaze information to help the model obtain human cognition general features. (c) The Gaze Balance Loss (GBL) helps the model solve the over/under-segmentation problem. } \label{Methods}
\vspace{-0.2cm}
\end{figure*}

\begin{figure}[t]
\centering
\includegraphics[width=0.83\columnwidth]{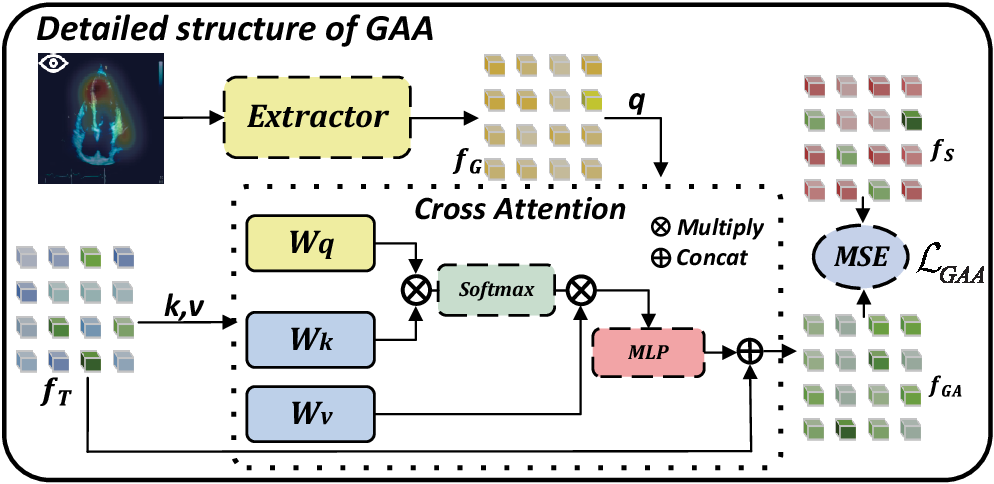}
\vspace{-0.25cm}
\caption{The GAA helps the student model leverage general features to acquire the ability to recognize and segment regions across domains.} \label{GAA}
\vspace{-0.2cm}
\end{figure}

The Gaze Augment Align Module (GAA) allows the student to learn the human cognition general features of the source domain and target domain. The previous feature alignment module\cite{feng2023unsupervised} ignore the difference of features and human recognition patterns in feature spaces. Human cognition general features help student model acquire human cross-domain recognition capabilities. As shown in Fig. \ref{Methods} (b), using the gaze heatmap help the framework obtain human cognition general features and get rid of the interference of source-specific features. As shown in Fig.\ref{GAA}, we extract human cross-domain recognition information from the gaze heatmap using the extractor to acquire the human cross-domain cognition general featrues ${f}_{G}$. Then, we obtain the feature matrix ${f}_{T}$ from the teacher model, which serves as the embeddings input $k$ (key) and $v$ (value) for the cross attention\cite{dosovitskiy2020ViT}, while ${f}_{G}$ serves as the input $q$ (query). We use the self-attention to integrate the gaze feature ${f}_{G}$ into ${f}_{T}$, extracting human cognition general features ${f}_{GA}$ shared by source domain and target domain. The GAA formula is as follows:
\begin{equation}
\small
    {f}_{GA}=Concat({f}_{T},Softmax({f}_{G}{f}_{T}/\sqrt{d}){f}_{T})
\label{GAA1}
\end{equation}
The cross-attention ensures that the human-centric cross-domain recognition information of the gaze heatmap is preserved. The alignment loss ${\mathcal L}_{GAA}$ is achieved by comparing the human cognition general features ${f}_{GA}$ with the ${f}_{S}$ of the student with MSE loss.

\subsection{Gaze Balance Loss}
The human-centric gaze heatmaps are able to help the student address the issues of under/over-segmentation when the training is unsupervised. Previously, the training effectiveness of the model in the target domain depends on the quality of pseudo-labels. To help the model address the overfitting caused by incomplete pseudo-labels in domain adaptation, we proposed a human cognition guided loss called that gaze balance loss(GBL). Doctors focus on the difficult-to-segment areas of the cardiac ultrasound images. The human-centric gaze heatmap provide guidance that incorporate human cognition guided knowledge for cross-entropy without labels. 
Using gaze heatmaps as weights to adjust the loss in under/over-segmentation areas is able to eliminate the domain gap and reduce the interference from incomplete pseudo-label. As shown in Fig. \ref{Methods} (c), heat regions indicate areas that receive more attention from doctors. We generate a weight mask $w$ from the regularized gaze heatmap. After obtaining the $w$, we multiply it with the output results $\hat{Y}_{t}^{S}$ of the student model to obtain the weighted output results. Then, we calculate the cross-entropy loss between the weighted output results and the pseudo labels to obtain the ${L}_{GB}$ in eq.(\ref{GB}):
\begin{equation}
\label{GB}
\small
    {\mathcal L}_{GB}=-\frac{1}{N}\displaystyle\sum_{i=1}^{N}[{y}_{i}log({w}_{i}\hat{y}_{i})+(1-{y}_{i})log(1-{w}_{i}\hat{y}_{i})]
\end{equation}

\begin{figure}[t]
\centering
\includegraphics[width=\columnwidth]{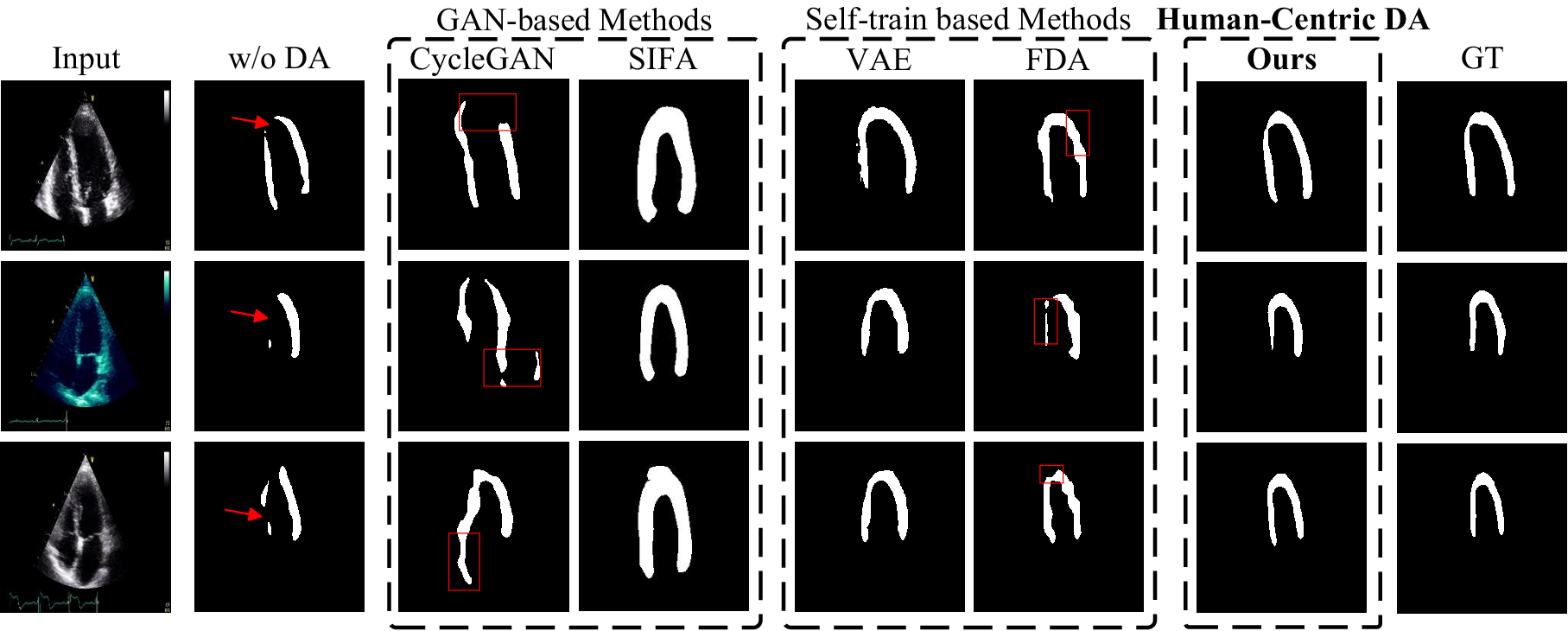}
\vspace{-0.7cm}
\caption{Visual comparison of outputs. The results show that our GAHCDA uses gaze information to obtain results closer to GT.} \label{fig3}
\vspace{-0.2cm}
\end{figure}

\subsection{Training Strategy}
We train the teacher model using the source domain data to obtain parameters $\theta_{t}$. The student model parameters $\theta_{s}$ initialized with $\theta_{t}$. After obtaining ${L}_{GAA}$ and ${L}_{GB}$, We calculate the loss using the formula in eq.(\ref{total}):
\begin{equation}
\label{total}
\small
    \mathop{\arg\min}\limits_{\theta_{s}} (\lambda_{gaa}{\mathcal L}_{GAA}+\lambda_{gb}{\mathcal L}_{GB}+ \lambda_{dice}{\mathcal L}_{DICE}+\lambda_{ce}{\mathcal L}_{CE})
\end{equation}
${\mathcal L}_{DICE}$ represents the DICE loss, and ${\mathcal L}_{CE}$ represents the cross-entropy loss with inputs $\hat{Y}_{t}^{S}$ and ${Y}_{t}$. We use $\lambda_{gaa}$, $\lambda_{gb}$, $\lambda_{dice}$ and $\lambda_{ce}$ to control the weights of different losses and achieve better output results.
\section{Experiments}
\subsection{Experiment Setting}
Our experiment used publicly available datasets CAMUS\cite{leclerc2019deep} as source domain with 9964 cardiac ultrasound images, and HMC-QU\cite{degerli2024early} as target domain with 2349 images.
%CAMUS includes 9964 images cardiac ultrasound images, and HMC-QU . 
A Tobii Eye-Tracker 5 record the doctor's gaze. %trajectories. 
%The eye tracker samples the coordinates of the doctor's gaze at a rate of five times per second.
The eye tracker samples coordinates of the gaze with five times per second.
Our model was implemented in PyTorch 1.13.0 and trained with NVIDIA RTX4070Ti GPU. We used U-Net as the backbone of the teacher/student model. The epoch is set to 200, and the batch size is 32. Model is optimized using RMSProp, with an initial learning rate of $1.0\times{10}^{-5}$. DSC (Dice similarity coefficient, \%) and ASSD (Average symmetric surface distance) are used to evaluate model. 

\renewcommand{\arraystretch}{1.2}
\begin{table}[tb]
\caption{The quantitative evaluation}\label{tab1}
\vspace{-0.25cm}
\centering
\begin{tabular}{l|l|c|c}
\hline
Strategy                           & Method         & DSC$_{\pm std}$$\uparrow$     & ASSD$_{\pm std}$$\downarrow$  \\  \hline
Upper bound                              & Supervised     & 90.23$_{\pm 2.3}$ & 4.098$_{\pm 0.9}$   \\ 
Lower bound                             & w/o DA & 70.14$_{\pm 4.5}$ & 10.28$_{\pm 2.1}$   \\ \cline{1-2}
\multirow{2}{*}{GAN-based DA}         & SIFA\cite{chen2019synergistic}   & 46.14$_{\pm 6.7}$ & 7.209$_{\pm 3.2}$   \\  
                                   & CycleGAN\cite{CycleGAN2017}       & 43.49$_{\pm 2.3}$ & 12.90$_{\pm 0.2}$    \\ \cline{1-2} 
\multirow{2}{*}{Self-train based DA} & VAE\cite{yaoVAE2022} & 54.16$_{\pm 5.4}$ & 10.59$_{\pm 2.0}$   \\  
                                   & FDA\cite{yang2020fda}            & 71.05$_{\pm 3.6}$ & 8.842$_{\pm 1.5}$  \\ \cline{1-2}
    \textbf{Human-centric DA}   &\textbf{Ours}            & \textbf{76.14$_{\pm 3.2}$} & \textbf{6.976$_{\pm 1.4}$}    \\ \hline
\end{tabular}
\vspace{-0.25cm}
\end{table}

\subsection{Comparison Study}
\subsubsection{Quantitative Evaluation}
The results shows that our GAHCDA is able to eliminate the domain gap between source and target domain effectively with human cognition. As the results shown in Table \ref{tab1}, our GAHCDA achieves the best segmentation results compared with other methods with DSC of 76.14\%, and ASSD of 6.976. Compared with SIFA, our method achieves 30\% DSC improvements. Compared with the FDA, our method achieves 5.09\% DSC improvements. The experimental results show that our GAHCDA method attains state-of-the-art performance in the domain adaptation of cardiac ultrasound image segmentation. 
\subsubsection{Qualitative Evaluation}
GAHCDA extract the structure and cardiac texture of the target by leveraging human cognition, resulting in outputs that are closer to the ground truth (GT). As shown in Fig. \ref{fig3}, w/o DA, the model shows under-segmentation in the target domain. The results of GAN-based methods show over/under-segmentation occurred in the area within the red box. Self-train based methods are affected by low-quality pseudo-labels, which fail to eliminate source-specific features, resulting in incomplete segmentation. The output mask of VAE has a large gap with GT in morphology which shows the instability of VAE. The FDA shows distortions in red box and under-segmentation which shows that the gap between domains is not only reflected in the frequency domain. Due to the effectiveness of GAA and GBL, the results of GAHCDA exhibit the best performance in terms of morphology and continuity.

\subsection{Ablation Study}
\subsubsection{Ablation Study}
\begin{figure}[t]
\centering
\includegraphics[width=\columnwidth]{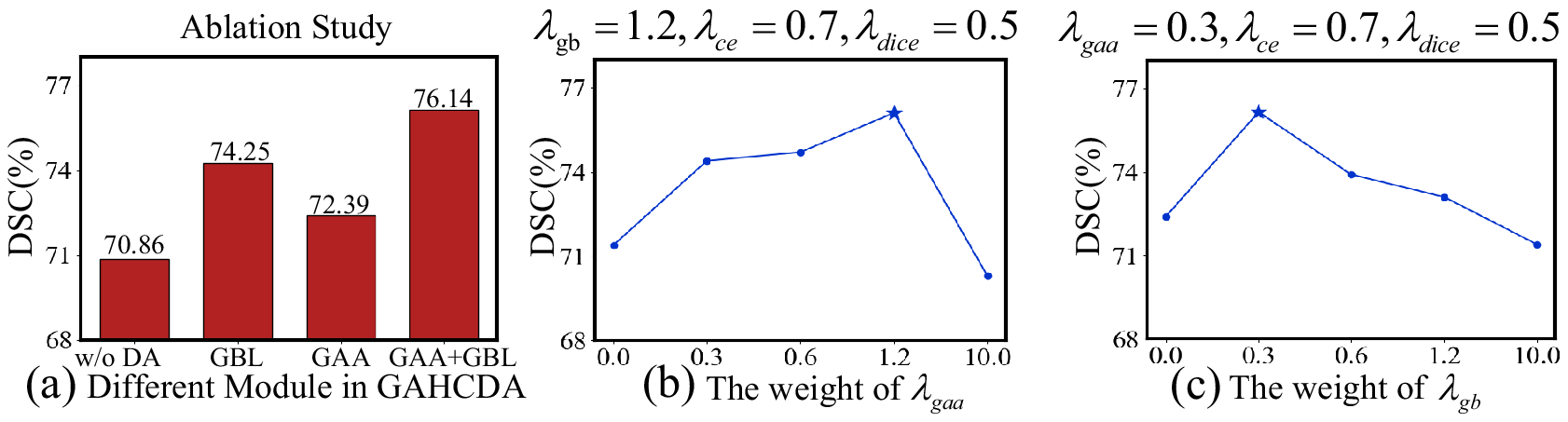}
\vspace{-0.6cm}
\caption{The ablation study demonstrated that GAA and GBL effectively eliminate domain differences by leveraging human cognition, thereby improving the model's segmentation performance in the target domain. Ablation study of (a) GAHCDA. (b) Hyper-parameter $\lambda_{gaa}$. (c) Hyper-parameter $\lambda_{gb}$.} \label{Ablation_Study}
\end{figure}

As shown in Fig. \ref{Ablation_Study} (a), we validated the effectiveness of the GAA and GBL. GAA improved the DSC by 1.53\%, while GBL showed an improvement compared to no adaptation, with a 3.39\% increase in DSC. The full module combines the advantages of GAA and GBL, obtaining human cognition general features while achieving more complete segmentation in the gaze area. The DSC increased by 5.28\% compared to no adaptation. 

\begin{figure}[t]
\centering
\includegraphics[width=\columnwidth]{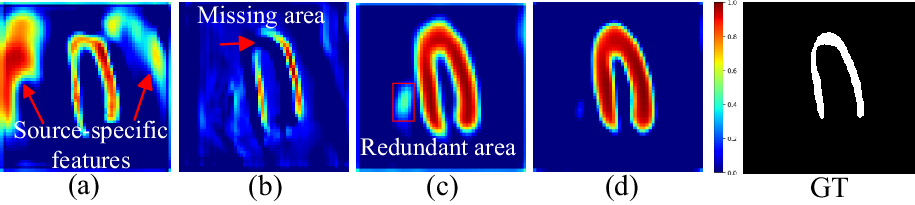}
\vspace{-0.6cm}
\caption{The feature map shows GAA and GBL help students reduce the inference of domain-specific features and obtain the ability of human cross-domain recognition. The feature map of (a) w/o DA, (b) with GAA, (c) with GBL, (d) with GAA\&GBL. } \label{Model_analysis}
\end{figure}

\begin{figure}[t]
    \centering
    \includegraphics[width=0.87\columnwidth]{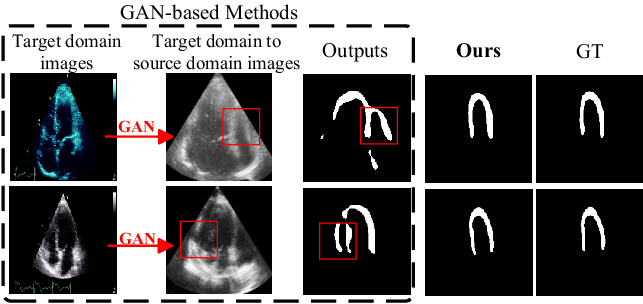}
    \vspace{-0.2cm}
    \caption{Model Analysis of GAN-based Methods. Without the support of human cross-domain recognition, GAN-based methods retain an excess of source-specific features, which caused interference to the student model.The red box shows the segmentation affected by mode collapse.}
    \label{fig_gan}
\end{figure}
\begin{figure}[t]
    \centering
    \includegraphics[width=0.6\columnwidth]{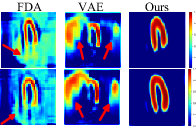}
    \vspace{-0.2cm}
    \caption{Model Analysis of self-train based methods. Without the support of human cross-domain recognition guidance module GAA and GBL, self-train based methods shows a redundant source-specific features interference on target domain affected by the pseudo label. }
    \label{fig_selftrain}
\end{figure}
\subsubsection{Hyper-parameter Analysis}
 As shown in Fig. \ref{Ablation_Study} (b)\&(c), as these two hyper-parameters are gradually increased, the performance of our framework on the target domain continuously improves. Increasing the weight of $\lambda_{gb}$ enhances the attention of the GBL to the over/under-segmentation areas in the output mask with the human cognition.
Also, GAA strengthens the role of GBL and increases the attention of target areas, GAA ensures the consistency of features and GBL optimization lead to better performance in the target domain.
\subsection{Model Analysis}
\subsubsection{Analysis of GAHCDA}
We extracted the feature maps of decoder with different modules. As shown in Fig. \ref{Model_analysis} (a), w/o DA, feature redundancy appears in the feature map. As shown In Fig. \ref{Model_analysis} (b), with GAA, the module extracts cross-domain human cognition general features guided by the gaze map which help the student model eliminate the domain gap between the source and target domain. But from the feature map, we can see the segmentation area features are still incomplete. In Fig. \ref{Model_analysis} (c), with GBL, we can see features are complete, but without GAA, there are redundant features in red box, the segmentation output on the target domain is interfered with redundant features. As shown in Fig. \ref{Model_analysis} (d), the full GAHCDA model has a complete and clear feature map. 
\subsubsection{Analysis of GAN-Based DA}
Due to the domain gap in different domain cardiac ultrasound images, GAN-based DA failed to adapt to target domain. As shown in Fig. \ref{fig_gan}, target to source images retain a amount of source-specific features, which disrupt the original structure of the target domain images. In the first row of Fig. \ref{fig_gan}, the adapted target image shows an extra region in the red box, which negatively affects the segmentation results. In the second row, the adapted target image also exhibits a redundant region, and this redundant region is output in the output mask. 
\subsubsection{Analysis of Self-Train Based DA}
GAHCDA can eliminate the interference of source-specific features. The previous self-train based methods, due to the lack of human cognitive guidance, fail to enable the student model to distill the correct cross-domain feature recognition patterns from the teacher model. Without the help of cognitive guidance loss, these methods are prone to feature redundancy. As shown in Fig. \ref{fig_selftrain}, the features of FDA and VAE exhibit a significant amount of redundancy, which is related to the source domain. The source-specific features lead to over/under-segmentation. Our GAHCDA leverages human cognitive guidance to eliminate the interference of source-specific features, resulting in more complete and clear segmentation outcomes.
\section{Conclusion}
In this paper, we propose a gaze-assisted human-centric domain adaptation framework for cardiac ultrasound segmentation. The model innovatively proposes the following modules: (1) Gaze Augment Align Module: Using gaze information combined with the cross-attention module to help student model learn human cognition general features in the source domain and target domain to improve performance of student. (2) Gaze Balance Loss: Use the gaze heatmaps to give more weight to the discontinuous areas. The ablation study proves that the gaze augment align module and gaze balance loss are helpful to the model in the domain adaptation.
\bibliographystyle{IEEEtran}
\bibliography{IEEEabrv,ref}

\end{document}